\documentclass[letterpaper]{article} % DO NOT CHANGE THIS
\usepackage{aaai25}  % DO NOT CHANGE THIS
\usepackage{times}  % DO NOT CHANGE THIS
\usepackage{helvet}  % DO NOT CHANGE THIS
\usepackage{courier}  % DO NOT CHANGE THIS
\usepackage[hyphens]{url}  % DO NOT CHANGE THIS
\usepackage{graphicx} % DO NOT CHANGE THIS
\usepackage{natbib}  % DO NOT CHANGE THIS AND DO NOT ADD ANY OPTIONS TO IT
\usepackage{caption} % DO NOT CHANGE THIS AND DO NOT ADD ANY OPTIONS TO IT
\usepackage{booktabs} % 提供表格优化
\usepackage{amsmath} % 必要的数学包
\usepackage{amssymb}
\usepackage{enumitem}
\usepackage{multirow}
\usepackage{pgfplots}
\usepackage{pgfplotstable}
\pgfplotsset{compat=newest}

\usepackage{algorithm}
\usepackage{algorithmic}

\usepackage{newfloat}
\usepackage{listings}
\DeclareCaptionStyle{ruled}{labelfont=normalfont,labelsep=colon,strut=off} % DO NOT CHANGE THIS
\floatstyle{ruled}
\newfloat{listing}{tb}{lst}{}
\floatname{listing}{Listing}
\title{A Generalizable Anomaly Detection Method in Dynamic Graphs}
\author {
    Xiao Yang$^{1,2,3}$,
    Xuejiao Zhao$^{1,2}$\thanks{Corresponding Author},
    Zhiqi Shen$^{1}$\footnotemark[1]
}
\affiliations {
    $^1$ College of Computing and Data Science, Nanyang Technological University (NTU), Singapore\\
    $^2$ Joint NTU-UBC Research Centre of Excellence in Active Living for the Elderly (LILY), NTU, Singapore\\
    $^3$ Joint NTU-WeBank Research Centre on Fintech, NTU, Singapore\\
    yangxiao.cs@gmail.com, \{xjzhao,zqshen\}@ntu.edu.sg
}

\usepackage{bibentry}
\begin{document}

\maketitle

\begin{abstract}
Anomaly detection aims to identify deviations from normal patterns within data. This task is particularly crucial in dynamic graphs, which are common in applications like social networks and cybersecurity, due to their evolving structures and complex relationships. Although recent deep learning-based methods have shown promising results in anomaly detection on dynamic graphs, they often lack of generalizability. 
In this study, we propose GeneralDyG, a method that samples temporal ego-graphs and sequentially extracts structural and temporal features to address the three key challenges in achieving generalizability: Data Diversity, Dynamic Feature Capture, and Computational Cost.  Extensive experimental results demonstrate that our proposed GeneralDyG significantly outperforms state-of-the-art methods on four real-world datasets.
\end{abstract}

\begin{links}
    \link{Code}{https://github.com/YXNTU/GeneralDyG}
    % \link{Datasets}{https://aaai.org/example/datasets}
    % \link{Extended version}{https://aaai.org/example/extended-version}
\end{links}

\section{Introduction}
Graphs are extensively employed to model complex systems across various domains, such as social networks~\cite{wang2019influence}, human knowledge networks~\cite{ji2021survey}, e-commerce~\cite{qu2020category}, and cybersecurity~\cite{gao2020hincti}. Although the bulk of researches focus on static graphs, real-world graph data often evolves over time~\cite{skarding2021foundations}. Taking knowledge networks as an example, there is new knowledge being added to the network every month, with connections between different concepts evolving over time. To model and analyze graphs where nodes and
 edges change over time, mining dynamic graphs gains
 increasing popularity in the graph analysis. Anomaly detection in dynamic graphs~\cite{ma2021comprehensive,ho2024graphanomalydetectiontime} is vital for identifying outliers that significantly deviate from normal patterns such as anomalous edges or anomalous nodes,including the detection of fraudulent transactions, social media spam, and network intrusions~\cite{dou2020enhancing}. By utilizing the temporal information and relational structures inherent in dynamic graphs, researchers can more effectively identify anomalies, thereby enhancing the security and integrity of various systems~\cite{pourhabibi2020fraud}.

Recently, techniques based on deep learning have facilitated significant advancements in anomaly detection within dynamic graphs. For example, methods like GDN~\cite{deng2021graph}, StrGNN~\cite{cai2021structural} focus on extracting structural information from graphs, while approaches such as LSTM-VAE~\cite{park2018multimodal} and TADDY~\cite{liu2021anomaly} concentrate on capturing temporal information.In addition, self-supervised~\cite{lee2024slade} and semi-supervised~\cite{tian2023sad} methods have also been applied to dynamic graph anomaly detection.

Despite their improved performance, current deep learning-based methods lack the crucial generalizability~\cite{brennan1992generalizability} needed for dynamic graph tasks across different tasks or datasets. A model with strong generalization can adapt to different tasks without significant adjustments to its architecture or parameters, reducing the need for retraining or redesigning for new tasks\cite{bai2022temporal}. Conversely, in anomaly detection, where identifying potential risks or issues is crucial, poor generalization may lead to missed critical anomalies in new scenarios, thereby diminishing the model's reliability in real-world applications. Specifically, the inadequate encoding of anomalous events\footnote{In this paper, both node anomalies and edge anomalies are collectively referred to as anomalous events.} in existing methods results in poor generalization. Firstly, in the absence of raw event attributes, they fail to generate informative event encodings that accurately represent the properties of the events. For example, SimpleDyG~\cite{wu2024feasibility} nearly discards all topological structure information, tokenizing only the nodes while ignoring the edges, which leads to the loss of critical structural information during node prediction tasks, making it unsuitable for node anomaly detection tasks and even less so for edge anomaly detection. The positional encoding method in TADDY~\cite{liu2021anomaly} may not capture structural similarities and could fail to model the structural interactions between events, as demonstrated in SAT~\cite{chen2022structure}. TADDY's node position-specific encoding may result in ambiguous structural information, leading to suboptimal results in node anomaly detection tasks. Furthermore, some methods, such as GDN~\cite{deng2021graph}, exhibit inadequate temporal information capture capabilities. For instance, GDN does not incorporate the information provided by specific time values when modeling temporal data, resulting in poor performance on time-sensitive datasets such as Bitcoin-Alpha and Bitcoin-OTC~\cite{liu2021anomaly}. 

Developing a highly generalizable dynamic graph anomaly detection method presents several challenges, primarily in: 1. \textbf{Data Diversity}: Differences across dynamic graph datasets, such as topological structures and node and edge attributes, can be substantial. The method must identify and adapt to a wide range of feature distributions. 2. \textbf{Dynamic Feature Capture}: Anomalies in dynamic graphs may occur locally (e.g., anomalous behavior of specific nodes or edges) or globally (e.g., abnormal changes in network topology). The method must capture both local and global dynamic features. 3. \textbf{Computational Cost}: Dynamic graph anomaly detection often involves large-scale graph data, making computational resources and time efficiency significant challenges.

Hence, in this work, we propose a novel approach for anomaly detection named GeneralDyG, which addresses the three key challenges mentioned above and ensures generalizability in dynamic graph anomaly detection tasks. It ensures simplicity by sampling ego-graphs around anomalous events, then uses a novel GNN extractor to capture structural information, and finally employs a Transformer module to capture temporal information. Specifically, the main contributions of our work are:
\begin{itemize}
    \item We design a novel GNN extractor, which embeds nodes, edges, and topological structures into the feature space. By alternating the message-passing perspective between nodes and edges, it performs graph convolution on both simultaneously. This ensures that GeneralDyG adapts to diverse feature distributions.
    \item We introduce special tokens into the feature sequences to distinguish the hierarchical relationships between anomalous events, ensuring that the method captures global temporal information while maintaining focus on local dynamic features.
    \item We design a novel ego-graph\footnote{The subgraph that consists of the ego events and all events within the $k$-hop range from the ego event.} sampling method for training anomalous events instead of using the entire graph. This approach significantly reduces computational resources, enhancing the overall efficiency of the method.
    \item We demonstrate the effectiveness of GeneralDyG on four benchmark datasets for detecting anomalous events, showing that it achieves better performance than state-of-the-art anomaly detection methods.
\end{itemize}

\section{Related Work}
\subsection{Anomaly Detection in Dynamic Graphs}
Anomalies are infrequent observations that significantly deviate from the rest of the sample, such as data records or events. Dynamic graph anomaly detection primarily focuses on identifying unusual events within a dynamic graph~\cite{Ekle_2024, ho2024graphanomalydetectiontime, ma2021comprehensive}. Recently, deep learning methods have made significant advancements in anomaly detection for dynamic graphs. Modeling time series-related tasks as anomalous node detection in dynamic graphs is considered a viable approach~\cite{su2019robust, chen2022deep, zhang2022grelen, dai2022graph}. Specifically, M-GAT employs a multi-head attention mechanism along with two relational attention modules—namely, intra-modal and inter-modal attention—to explicitly model correlations between different modalities~\cite{ding2023mst}. MTAD-GAT incorporates two parallel graph attention layers to capture the complex dependencies in multivariate time series across both temporal and feature dimensions~\cite{zhao2020multivariate}. GDN integrates structural learning with graph neural networks and leverages attention weights to enhance the explainability of detected anomalies~\cite{deng2021graph}. FuSAGNet optimizes reconstruction and forecasting by combining a Sparse Autoencoder with a Graph Neural Network to model multivariate time series relationships and predict future behaviors~\cite{han2022learning}.

Detection of edge anomalies in dynamic graphs has also garnered increasing attention. Classical methods include the randomized algorithm SEDANSPOT~\cite{eswaran2018sedanspot} and the hypothesis-based approach Midas~\cite{bhatia2020midas}. Many recent methods have employed discrete approaches to address this task. For instance, Addgraph utilizes a GCN to extract graph structural information from slices, followed by GRU-attention~\cite{zheng2019addgraph}. StrGNN extracts $h$-hop closed subgraphs centered on edges and employs GCN to model structural information on snapshots, with GRU capturing correlations between snapshots~\cite{cai2021structural}. Recently, SAD introduced a continuous dynamic approach for anomaly detection using a semi-supervised method~\cite{tian2023sad}.

\subsection{Transformer on Dynamic Graphs}
Transformers are a type of neural network that rely exclusively on attention mechanisms to learn representative embeddings for various types of data, as initially introduced in~\cite{vaswani2017attention}. Recent works have also applied Transformers to dynamic graph tasks. For instance, GraphERT pioneers the use of Transformers to seamlessly integrate graph structure learning with temporal analysis by employing a masked language model on sequences of graph random walks~\cite{beladev2023graphert}. GraphLSTA captures the evolution patterns of dynamic graphs by effectively extracting and integrating both long-term and short-term temporal features through a recurrent attention mechanism~\cite{gao2023anomaly}. Taddy employs a Transformer to handle diffusion-based spatial encoding, distance-based spatial encoding, and relative time encoding, subsequently deriving edge representations through a pooling layer to calculate anomaly scores~\cite{liu2021anomaly}. SimpleDyG reinterprets dynamic graphs as a sequence modeling problem and presents an innovative temporal alignment technique. This approach not only captures the intrinsic temporal evolution patterns of dynamic graphs but also simplifies their modeling process~\cite{wu2024feasibility}.

\section{Preliminaries}
\textbf{Notations. } A continuous-time dynamic graph (CTDG) is used to represent relational data in evolving systems. A CTDG is defined as $\mathcal{G} = (\mathcal{V}, \mathcal{E})$, where $\mathcal{V}$ is the set of nodes that participate in temporal edges, and $\mathcal{E}$ is a chronologically ordered series of edges. Each edge $\delta(t) = (v_i, v_j, t, e_{ij})$ represents an interaction from node $v_i$ to node $v_j$ at time $t$ with an associated feature $e_{ij}$. The node attributes for nodes $v_i, v_j \in \mathcal{V}$ are denoted by $x_{v_i}, x_{v_j} \in \mathbb{R}^d$, and the node attributes for all nodes are stored in $\mathcal{X} \in \mathbb{R}^{n \times d}$. Additionally, the edge attributes for edges $e_{ij} \in \mathcal{E}$ are denoted by $y_{e_{ij}} \in \mathbb{R}^d$, and the edge attributes for all edges are stored in $\mathcal{Y} \in \mathbb{R}^{m \times d}$, where $n$ is the number of nodes and $m$ is the number of edges in the CTDG. In this paper, we explore a method called GeneralDyG for handling node-level and edge-level anomalies. Therefore, in the following text, we treat nodes $\mathcal{V}$ and edges $\mathcal{E}$ collectively as anomaly events $\mathcal{A}$. Similarly, we consider node features $\mathcal{X}$ and edge features $\mathcal{Y}$ together as anomaly features $\mathcal{Z}$.

\noindent \textbf{Transformer on  CTDG.} While Graph Neural Networks (GNNs) directly leverage the inherent structure of graphs, Transformers take a different approach by inferring relationships between nodes using their attributes rather than the explicit graph structure~\cite{dwivedi2020generalization}. Transformer treats the dynamic graph as a collection of edges, utilizing the self-attention mechanism to identify similarities between them. The architecture of the Transformer\cite{fang2023annotations, fang2023hierarchical} consists of two fundamental components: a self-attention module and a feed-forward neural network.

In the self-attention module, the input anomaly features $\mathcal{Z}$ are initially projected onto the query ($Q$), key ($K$), and value ($V$) matrices through linear transformations, such that $Q = \mathcal{Z} W_Q$, $K = \mathcal{Z} W_K$, and $V = \mathcal{Z} W_V$, respectively. The self-attention can then be computed as follows:
\begin{equation} \label{QKV}
\text{Attn} (\mathcal{Z}) = \text{softmax}\left(\frac{QK^T}{\sqrt{d_{\text{out}}}}\right)V \in \mathbb{R}^{(m+n) \times d_{\text{out}}}.
\end{equation}

To address dynamic graph tasks, multiple Transformer layers can be stacked to build a model that provides node-level representations of the graph~\cite{wang2021tcl}. However, due to the permutation invariance of the self-attention mechanism, the Transformer generates identical representations for nodes with the same attributes, regardless of their positions or surrounding structures within the graph. This characteristic necessitates the incorporation of positional and contextual information into the Transformer, typically achieved through positional encoding~\cite{cong2021spatial, sun2022dual}.

\noindent \textbf{Absolute encoding.} Absolute encoding involves adding or concatenating positional or structural representations of the graph to the input node features before feeding them into the main Transformer model. Examples of such encoding methods include Laplacian positional encoding~\cite{dwivedi2020generalization}, Random Walk Positional Encoding~\cite{dwivedi2021graph}, and Node Encoding~\cite{liu2021anomaly}. A key limitation of these methods is that they typically fail to capture the structural similarity between nodes and their neighborhoods, thereby not effectively leveraging the graph's structural information.

\noindent \textbf{Problem Definition.} The goal of this paper is to detect anomalous edges and nodes at each timestamp. Based on the previously mentioned notations, we model anomaly detection in dynamic graphs as a task of computing anomaly scores.

\noindent \textit{Definition 1.} Given a dynamic graph $\mathcal{G}$, where each $\mathcal{G}_t = (\mathcal{V}_t, \mathcal{E}_t)$ represents the graph at timestamp $t$, the goal of anomaly detection is to identify unusual edges and nodes within this evolving structure. For each edge $e \in \mathcal{E}_t$ and each node $v \in \mathcal{V}_t$, the objective is to compute an anomaly score $f(e)$ and $f(v)$, respectively, where $f$ is a learnable anomaly score function. The anomaly score quantifies the degree of abnormality for both edges and nodes, with a higher score $f(e)$ or $f(v)$ indicating a greater likelihood of anomaly for edge $e$ or node $v$.

Building on previous research, we adopt an unsupervised approach for anomaly detection in dynamic graphs. During training, all edges and nodes are considered normal. Binary labels indicating anomalies are provided during the testing process to assess the performance of the algorithms. Specifically, a label $y_e = 1$ signifies that $e$ is anomalous, whereas $y_e = 0$ denotes that $e$ is normal. Similarly, a label $y_n = 1$ indicates that a node is anomalous. It is important to note that anomaly labels are often imbalanced, with the number of normal edges and nodes typically being much greater than the number of anomalous ones.
\begin{figure*}[h]
    \centering
    \includegraphics[width=1\linewidth]{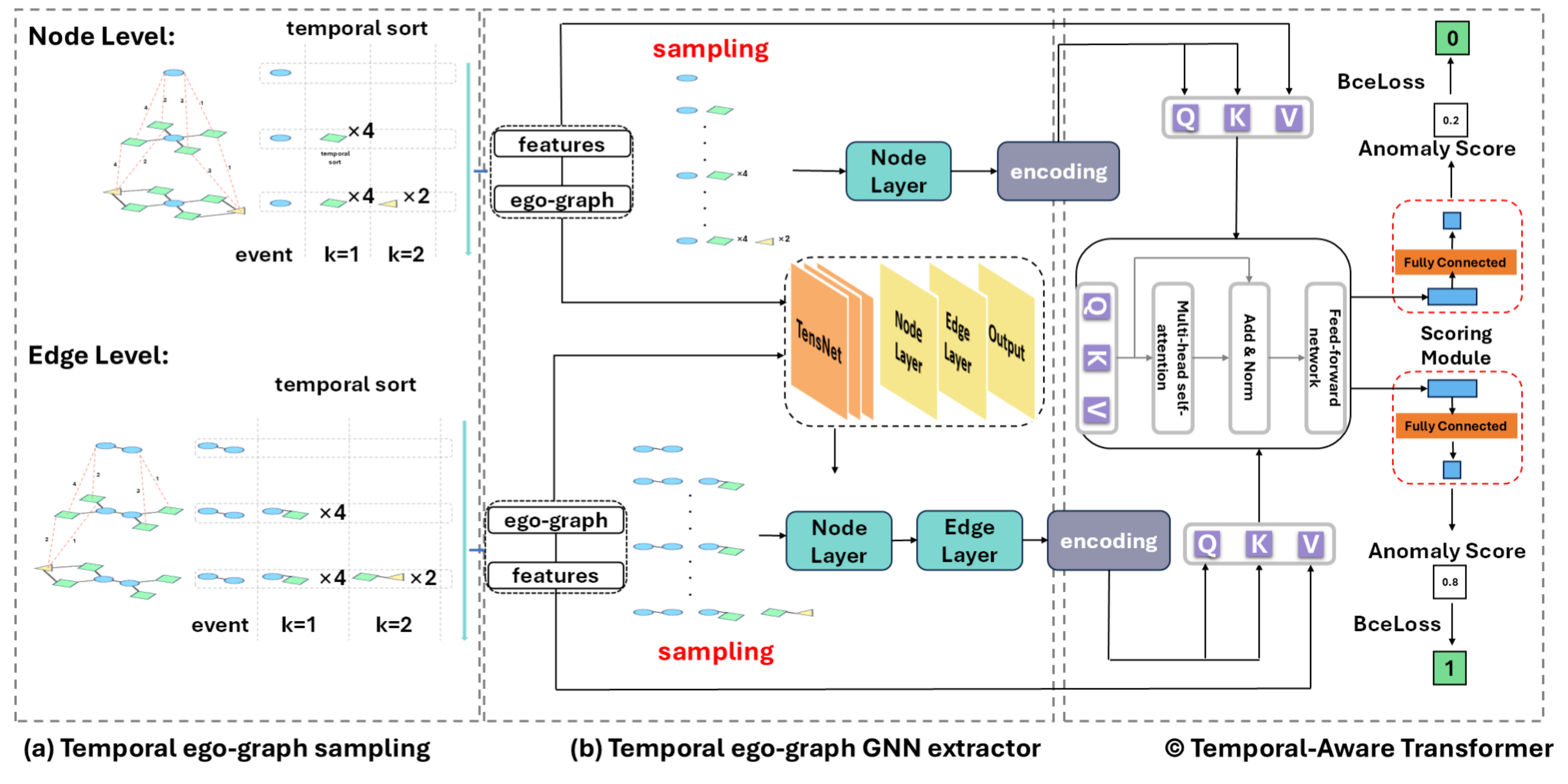}
    \caption{The proposed generalizable anomaly detection framework. GeneralDyG consists of three main components: (a)Temporal ego-graph sampling. (b)Temporal ego-graph GNN extractor. (c)Temporal-Aware Transformer}
    \label{process}
\end{figure*}
\section{Methodology}
In this section, we introduce the general framework of our approach, which consists of three main components: Temporal ego-graph sampling, Temporal ego-graph GNN extractor, and Temporal-Aware Transformer. An overview of the proposed framework is illustrated in Figure \ref{process}. Initially, we extract ego-graphs at the level of each anomaly event, capturing $k$-hop temporal dynamics. These temporal ego-graphs are then transformed into anomaly feature sequences, preserving their temporal and structural order, as demonstrated in Figure \ref{process}(a). To fully understand the structural information of these sequences, they are processed through a GNN model to extract the structural details of the temporal ego-graphs, as depicted in Figure \ref{process}(b). Finally, both the original sequence features and the structure-enriched sequence features are fed into the Transformer to evaluate the anomaly detection task, as shown in Figure \ref{process}(c).

\subsection{Temporal ego-graph sampling}
Unlike conventional methods that map dynamic graphs into a series of snapshots to obtain tokens, we use a more lightweight approach by employing anomalous events as tokens for the Transformer. Additionally, to acquire the contextual representation and hierarchical information of these anomalous events, we extract the temporal $k$-hop ego-graph of each event to capture historical interaction information across different structures.

Specifically, we denote $a_i \in \mathcal{A}$ as an event in $\mathcal{G}$. For each event $a_i$, we utilize a $k$-hop algorithm to extract the historically interacted events and construct a series of $k$-hop ego-graphs centered around $a_i$, representing subsets of the largest $k$-hop ego-graph. Explicitly, we denote the temporal $k$-hop ego-graph for $a_i$ as a chronologically ordered sequence $w_i = sampling(\langle a_i^1 \rangle, \langle a_i^1, a_i^2, a_i^3 \rangle, \ldots, \langle a_i^1, a_i^2, a_i^3, \ldots, a_i^{|w_i|} \rangle)$, where $|w_i|$ is the number of previously interacting events, and the maximum value of $|w_i|$ is the total number of events that have interacted with $a_i$. Note that $\forall 1 \leq j < j' \leq |w_i|$, $a_i^j$ and $a^{j'}i$ represent historical interactions $(a_i, a^j_i, e{i,j})$ and $(a_i, a^{j'}i, e{i,j'})$, respectively, such that $e_{i,j} \leq e_{i,j'}$.

When implementing feature sequences sorted by time, it is crucial to simultaneously consider the hierarchical information introduced by the $k$-hop algorithm. Specifically, for the central event $a_i$, the set of events $a_{i;k}^j$ extracted by the $k$-hop algorithm exhibits greater similarity compared to the set of events $a_{i;k+1}^j$ extracted by the ($k$+1)-hop algorithm, as they share the same shortest path to the central point $a_i$. To better capture this hierarchical information, we draw inspiration from natural language processing methods and add special tokens to the feature sequence. These tokens ensure that the event sets between two special tokens maintain a chronological order. During training, the Transformer module and GNN extractor can receive the following input:
\begin{align}
\text{input}_i &= \langle |KHS| \rangle, a_i, \langle |KHS| \rangle, a{i;1}^1, a_{i;1}^2, \ldots, a_{i;1}^{|a_{i;1}|}, \ldots \notag \\
&\quad \langle |KHS| \rangle, a_{i;k}^1, a_{i;k}^2, \ldots, a_{i;k}^{|a_{i;k}|}, \langle |KHS| \rangle,
\end{align}
where $\langle |KHS| \rangle$ is a special token signifying the beginning and end of the input hierarchical sequence. Specifically, adding such special tokens helps the model recognize and differentiate between the hierarchical layers of the ego-graph. It should be noted that we use sampled ego-graphs here to enhance the model's generalization capability. Therefore, the raw features obtained by the Transformer module and GNN extractor are denoted as $z_i$, where $z_i$ is a subset of $\text{input}_i$.
\subsection{Temporal ego-graph GNN extractor}
A practical approach to extracting local structural information at an event $a_i$ is to apply an existing GNN model to the input graph with event feature sequences $z_i$, and utilize the output representation at $a_i$ as the ego-graph representation $\varphi(z_i)$. It is important to highlight that, to showcase the flexibility of our model, the GNN model employed here should be both straightforward and capable of simultaneously processing node features $\mathcal{X}$ and edge features $\mathcal{Y}$. Formally, we denote the selected GNN model with $\mathcal{K}$ layers applied to $k$-hop ego-graphs $k$-DG as $\text{GNN}_k^{\mathcal{K}}$. The output representation $\varphi(z_i)$ can be expressed as:
\begin{equation}
\varphi(z_i) = \text{GNN}_k^{\mathcal{K}}(z_i).
\end{equation}

Next, we discuss the choice of the $\text{GNN}_k^{\mathcal{K}}$ model. When the dataset information is known prior to anomaly event prediction—such as in cases where the CTDG consists solely of node features—a conventional GNN model like GCN, GAT, or GIN can be effectively utilized to extract ego-graph structural information. However, for CTDGs with diverse attributes, including both node and edge features, we introduce the Temporal Edge-Node Based Structure Extractor GNN (TensGNN). TensGNN is specifically designed to accommodate more complex scenarios by concurrently processing both types of features.

TensGNN encodes events by alternately applying node and edge layers, thereby embedding events into a shared feature space. Specifically, TensGNN employs operations analogous to spectral graph convolution for message passing on events. The node Laplacian-adjacency matrix with self-loops is defined as:
\begin{equation}
\Bar{A_v} = D^{\frac{1}{2}}_v \left(A_v + I_v\right) D^{\frac{1}{2}}_v,
\end{equation}
where $D_v$ is the diagonal degree matrix of $A_v + I_v$, and $I_v$ is the identity matrix. The node-level propagation rule for node features in the $(K+1)$-th layer is defined as:
\begin{equation} \label{GNN_n}
H^{(K+1)}_v = \sigma \left(T^T H^{(K)}_e W'_e \odot \Bar{A_v} H^{(K)}_v W_v \right),
\end{equation}
where $\sigma$ represents the activation function, the matrix $T \in \mathbb{R}^{N_v \times N_e}$ is a binary transformation matrix, with $T_{ij}$ indicating whether edge $j$ connects to node $i$. The symbol $\odot$ represents the Hadamard product. $W'_e$ and $W_v$ are learnable parameters for edges and nodes, respectively. Similarly, the Laplacianized edge adjacency matrix is defined as:
\begin{equation}
\Bar{A_e} = D^{\frac{1}{2}}_e \left(A_e + I_e\right) D^{\frac{1}{2}}_e,
\end{equation}
where $D_e$ is the diagonal degree matrix of $A_e + I_e$, and $I_e$ is the identity matrix. The propagation rule for edge features is then defined as:
\begin{equation}
H^{(K+1)}_e = \sigma \left(T^T H^{(K)}_v W'_v \odot \Bar{A_e} H^{(K)}_e W_e \right).
\end{equation}

Here, the matrix $T$ is defined analogously to that in Equation \ref{GNN_n}, with $W'_v$ and $W_e$ representing the learnable weights for the nodes and edges, respectively. TensGNN alternates between stacking node layers and edge layers to iteratively refine the embeddings of both types of events. Specifically, to derive the final encoding of nodes, the last layer before the output is a node layer. Conversely, to obtain the final encoding of edges, the last layer before the output is an edge layer.

\subsection{Temporal-Aware Transformer}
To enhance the Transformer's understanding of the topological structure of the temporal ego-graph while preserving the original event features, we overlay the topological structure information onto the Query and Key, while retaining the original event features as the Value. This approach allows the model to leverage structural information for the attention mechanism while maintaining the integrity of the original feature values for effective representation. Formally, for the event feature to be predicted, $z_i \in \mathcal{Z}$, we adopt the method proposed in~\cite{mialon2021graphit} and rewrite the self-attention as kernel smoothing. The final embedding calculation is then given by:
\begin{equation}
\text{Attn}(z_i) = \sum_{z_j \in k-DG} \frac{\mathcal{F}_{\text{exp}}(z_i, z_j)}{\sum_{z_w \in k-DG} \mathcal{F}_{\text{exp}}(z_i, z_w)} \text{w}_V z_j,
\end{equation}
where $\text{w}_V$ is the linear value function of the original event feature $z_i$, and $\mathcal{F}{exp}$ is an exponential kernel (non-symmetric), parameterized by $\text{w}_{Q}$ and $\text{w}_{K}$:
\begin{align}
\mathcal{F}{exp}(x, x') &:= \text{exp}\left(\frac{\langle \text{w}_{Q} x , \text{w}_{K} x' \rangle}{\sqrt{d_{out}}}\right), \\
\text{w}_{V} &= \text{W} z_i + b \notag, \\
\text{w}_{Q} &= \text{W} \varphi(z_i) + b \notag, \\
\text{w}_{K} &= \text{W} \varphi(z_i) + b \notag,
\end{align}
where $\langle \cdot , \cdot \rangle$ denotes the dot product. By optimizing the objective function, we obtain the final embeddings for each anomaly feature $z_i$. These final embeddings are then fed into the scoring module to compute the anomaly scores. It is important to note that the scoring modules for node-level and edge-level anomalies differ in the datasets used in this paper. For edge-level anomalies, we directly use the final output embedding from the training process as the anomaly score. Conversely, for node-level anomalies, the final output consists of a set of binary labels indicating whether each time step is anomalous, which serves as the final anomaly score.

\section{Experiments}
\subsection{Experimental Setup}
\textbf{Datasets. } We use four real-world datasets, categorized into two types: Node-Level and Edge-Level anomaly detection tasks. For Node-Level, we utilize SWaT (Secure Water Treatment), a small-scale Cyber-Physical system managed by Singapore’s Public Utility Board, and WADI (Water Distribution), an extension of SWaT that includes a more extensive water distribution network. Both datasets provide data from normal operations and controlled attack scenarios to simulate real-world anomalies. For Edge-Level, we employ Bitcoin-Alpha and Bitcoin-OTC, which are trust networks of Bitcoin users trading on platforms from www.btc-alpha.com and www.bitcoin-otc.com, respectively. In these datasets, nodes represent users, and edges indicate trust ratings between them, capturing interactions and trust dynamics within the Bitcoin trading community.

\noindent \textbf{Experimental Design.} In our experiments, The settings for Bitcoin-Alpha and Bitcoin-OTC are identical to those used in TADDY~\cite{liu2021anomaly}. We inject anomalies into the test set at proportions of 1\%, 5\%, and 10\%. SWaT and WADI are identical to those used in GDN~\cite{deng2021graph}. AUC\footnote{https://en.wikipedia.org/wiki/AUC}, AP\footnote{https://builtin.com/articles/mean-average-precision} and F1\footnote{https://en.wikipedia.org/wiki/F-score} are used as the primary metrics to evaluate the performance of the proposed GeneralDyG and baselines. 

\noindent \textbf{Baselines.} We evaluated GeneralDyG against 20 advanced baselines, which are classified into two categories: graph embedding methods and anomaly detection methods. A detailed description of the baselines can be found in the Appendix.
\begin{table*}[h]
\centering
\begin{tabular}{c|cccccc|cccccc}
\hline
           & \multicolumn{6}{c|}{Bitcoin-Alpha}                                                                  & \multicolumn{6}{c}{Bitcoin-OTC}                                                                     \\ \cline{2-13} 
Methods    & \multicolumn{2}{c}{1\%}         & \multicolumn{2}{c}{5\%}         & \multicolumn{2}{c|}{10\%}       & \multicolumn{2}{c}{1\%}         & \multicolumn{2}{c}{5\%}         & \multicolumn{2}{c}{10\%}        \\ \cline{2-13} 
           & AUC            & AP             & AUC            & AP             & AUC            & AP             & AUC            & AP             & AUC            & AP             & AUC            & AP             \\ \hline
node2vec   & 69.10          & 9.17           & 68.02          & 7.31           & 67.85          & 9.95           & 69.51          & 8.31           & 68.83          & 6.45           & 67.45          & 4.77           \\
DeepWalk   & 69.85          & 8.56           & 68.74          & 9.68           & 67.93          & 10.78          & 74.23          & 10.58          & 73.56          & 9.41           & 72.87          & 8.22           \\ \hline
TGAT       & 85.32          & 11.36          & 84.16          & 11.08          & 83.98          & 12.05          & 88.87          & 16.87          & 87.59          & 15.24          & 87.55          & 15.37          \\
TGN        & 86.92          & 13.00          & 86.78          & 16.85          & 86.21          & 17.00          & 84.33          & 11.33          & 83.49          & 11.25          & 83.47          & 10.79          \\
ADDGRAPH   & 83.41          & 13.21          & 84.70          & 13.01          & 83.69          & 14.28          & 86.00          & 16.04          & 84.98          & 15.21          & 84.77          & 14.21          \\
StrGNN     & 85.74          & 12.56          & 86.67          & 13.99          & 86.27          & 14.68          & 90.12          & 18.34          & 87.75          & 18.68          & 88.36          & 18.10          \\
TADDY      & {\textbf{ 94.51}}    & 16.51          & \underline{93.41}         & 18.32          & \underline{94.23}          & 19.67          & {\underline{ 94.55}}    & 16.10          & \underline{93.40}         & 18.47          & \underline{94.25}          & 18.92          \\
SAD        & 90.69          & \underline{19.99}         & 90.55          & 21.08          & 90.33          & 22.99          & 91.88          & \underline{26.32 }         & 90.99          & \underline{27.33}          & 90.04          & \underline{26.79 }         \\
SLADE      & 90.32          & 18.78          & 89.99          & \underline{22.02}          & 88.71          & \underline{24.41}          & 91.53          & 20.32          & 91.24          & 22.11          & 91.01          & 20.04          \\ \hline
GeneralDyG    & \underline{94.01}         & \textbf{24.00} & \textbf{95.41}         & \textbf{24.02} & {\textbf{ 96.28}}    & \textbf{26.73} & \textbf{94.66} & {\textbf{ 27.89}}    & {\textbf{ 94.86}}    & \textbf{29.97} & {\textbf{ 95.59}}    & {\textbf{ 27.13}}    \\ \hline
\end{tabular}
\caption{ Anomaly detection performance comparison on Edge-Level datasets. The best performing method in each experiment is in bold and the second-best method is indicated with underlining.
}
\label{edge}
\end{table*}
\begin{itemize}
    \item \textbf{Graph Embedding Methods:} node2vec~\cite{grover2016node2vec}, DeepWalk~\cite{perozzi2014deepwalk}, TGAT~\cite{xu2020inductive}, TGN~\cite{rossi2020temporal}.
    \item \textbf{Anomaly Detection Methods:} ADDGRAPH~\cite{zheng2019addgraph}, StrGNN~\cite{cai2021structural}, TADDY~\cite{liu2021anomaly}, SAD~\cite{tian2023sad},  SLADE~\cite{lee2024slade}, PCA~\cite{shyu2003novel}, KNN~\cite{angiulli2002fast}, GDN~\cite{deng2021graph}, BTAD~\cite{ma2023btad}, GRN-100~\cite{tang2023gru}, DAGMM~\cite{zong2018deep}, MST-GAT~\cite{ding2023mst}, FuSAGNet~\cite{han2022learning}, LSTM-VAE~\cite{park2018multimodal}, MTAD-GAT~\cite{zhao2020multivariate}.
\end{itemize}

\subsection{Overall Performance}
\textbf{Edge Level. }
We compared our methods, GeneralDyG, with nine strong edge-level baseline methods, as shown in Table \ref{edge}. Our methods consistently outperformed the baselines across both Bitcoin-Alpha and Bitcoin-OTC datasets. The baselines, lacking sufficient structural or temporal information, did not achieve state-of-the-art results. Specifically, GeneralDyG demonstrated an average AUC improvement of approximately 3.2\% and 4.5\%, respectively, compared to the best-performing baseline on the Bitcoin-Alpha dataset. In terms of Average Precision (AP), GeneralDyG achieved a significant improvement, with up to 24\% in the 1\% anomaly detection setting, representing a 19.8\% increase over the best-performing baseline.

On the Bitcoin-OTC dataset, GeneralDyG also exhibited substantial gains, with an average AUC increase of about 3.6\% and an AP improvement of up to 20.2\% over the baselines. This demonstrates that our methods are more effective in generalizing and capturing the temporal dynamics necessary for robust anomaly detection in these datasets.
\begin{table}[h]
\centering
\scalebox{1}{ 
\begin{tabular}{c|c|c}
\hline
Methods  & SWaT           & WADI           \\ \hline
PCA      & 23.16          & 9.35           \\
KNN      & 7.83           & 7.75           \\ \hline
GDN      & 80.82          & 56.92          \\
BTAD     & 81.43          & 53.77          \\ 
GRN-100  & 74.96          & 48.28          \\
DAGMM    & 39.37          & 36.09          \\
MST-GAT  & 83.55          & 60.31          \\
FuSAGNet & {\underline{83.69} }    & \textbf{60.70} \\
LSTM-VAE & 73.85          & 24.82          \\
MTAD-GAT & 31.71          & 16.94          \\
\hline
GeneralDyG  & \textbf{85.19} & {\underline{60.43} }    \\ \hline
\end{tabular}
}
\caption{Anomaly detection F1 scoring comparison on Node-Level datasets. The best performing method in each experiment is in bold and the second-best method is indicated with underlining.}
\label{node}
\end{table}
\noindent \textbf{Node Level. } We compared our methods, GeneralDyG, with ten strong node-level baseline methods, as shown in Table \ref{node}. Our methods generally outperformed the baselines. Specifically, GeneralDyG achieved the highest F1 score on the SWaT dataset with 85.19\%, surpassing the second-best method, FuSAGNet, by 1.8\%. On the WADI dataset, GeneralDyG reached an F1 score of 60.43\%, which is slightly below FuSAGNet’s 60.70\%, but still demonstrates competitive performance.

The baselines, particularly those lacking robust temporal modeling capabilities like PCA and KNN, showed significantly lower F1 scores, with KNN performing the worst on both datasets. Compared to these methods, GeneralDyG shows a notable improvement of approximately 58\% on SWaT and 53\% on WADI in F1 score. Overall, these results highlight that our methods are better at generalizing and capturing the temporal dynamics necessary for effective anomaly detection in node-level datasets.
\subsection{Ablation Study}
\begin{table}[h]
\centering
\scalebox{1}{
\begin{tabular}{c|ccc}
\hline
\multirow{2}{*}{Method} & \multicolumn{2}{c}{Bitcoin-Alpha} & WADI           \\ \cline{2-4} 
                        & AUC             & AP              & F1             \\ \hline
GeneralDyG                      & \textbf{96.28}  & \textbf{26.73}  & \textbf{60.43} \\ \hline
w/o ego-graph           & 96.01           & 19.33           & 59.45          \\
w/o TensGNN             & 92.02           & 22.63           & 55.13          \\
w/o Transformer         & 93.71           & 20.20           & 58.46          \\ \hline
\end{tabular}}
\caption{
The performance of GeneralDyG and its variants on both Node-Level and Edge-Level datasets.
}
\label{ablation}
\end{table}
We conducted an ablation study to assess the contribution of each component in the proposed GeneralDyG, as detailed below:
\begin{itemize}
    \item \textbf{w/o ego-graph.} This variant omits the temporal ego-graph sampling process and directly uses the entire graph as input features.
    \item \textbf{w/o TensGNN.} This variant removes the GNN extractor, thereby omitting the extraction of structural information from the events.
    \item \textbf{w/o Transformer.} This variant excludes the Transformer module, thus omitting the extraction of temporal information from the events.
\end{itemize}
The ablation study results in Table \ref{ablation} highlight the significance of each component in GeneralDyG. Removing the temporal ego-graph sampling (w/o ego-graph) results in a decrease of AUC by 0.27\% and AP by 27.9\% on Bitcoin-Alpha, and a reduction in F1 score by 1.6\% on WADI, indicating its critical role in capturing temporal dependencies. Excluding the GNN extractor (w/o TensGNN) leads to a significant decrease in AUC by 4.26\%, AP by 15.8\% on Bitcoin-Alpha, and F1 score by 8.6\% on WADI, underscoring the importance of structural information. Removing the Transformer module (w/o Transformer) results in a decrease of AUC by 2.57\%, AP by 24.6\% on Bitcoin-Alpha, and F1 score by 3.3\% on WADI, emphasizing the need for temporal information processing. These results confirm that each component is crucial for achieving optimal performance.
\subsection{How to Set Up the Optimal GeneralDyG}
\begin{figure}[h]
    \centering
    \begin{tikzpicture}
        \begin{axis}[
            width=0.7\linewidth,
            height=0.6\linewidth,
            colorbar,
            colormap/hot2,
            xlabel={$k$},
            ylabel={$\mathcal{K}$},
            title={},
            ]
            \addplot [
                matrix plot*,
                point meta=explicit,
                mesh/cols=5
            ] table[meta=z] {
                x y z
                1 1 95.81
                2 1 95.87
                3 1 96.00
                4 1 96.13
                5 1 95.97
                1 2 96.18
                2 2 96.28
                3 2 96.11
                4 2 96.15
                5 2 96.08
                1 3 95.97
                2 3 95.93
                3 3 95.57
                4 3 95.24
                5 3 95.10
                1 4 95.01
                2 4 94.43
                3 4 94.88
                4 4 94.11
                5 4 94.01
                1 5 93.33
                2 5 93.15
                3 5 94.01
                4 5 93.20
                5 5 92.58
            };
        \end{axis}
    \end{tikzpicture}
    \caption{Effect of Parameters $k$ and $\mathcal{K}$ on Bitcoin-Alpha}
    \label{fig:heatmap}
\end{figure}
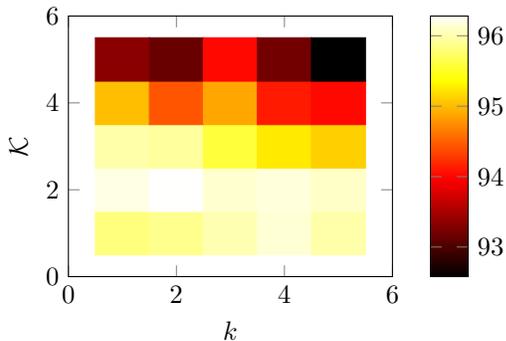
The heatmap in Figure~\ref{fig:heatmap} illustrates the impact of the parameters \(k\) and \(\mathcal{K}\) on model performance for the Bitcoin-Alpha dataset. It indicates that higher values of \(\mathcal{K}\) (the number of layers in the TensGNN) generally lead to decreased performance, suggesting that having too many layers can be detrimental to the model's effectiveness. This could be due to overfitting or increased model complexity without corresponding gains in performance.

On the other hand, the parameter \(k\) (which controls the temporal ego-graph sampling) has a less pronounced effect on performance. While increasing \(k\) does affect the results, it mainly impacts the training process due to the additional parameters it introduces. 

Thus, the optimal setup should aim for a balance: choosing a modest number of layers (\(\mathcal{K}\)) to avoid overfitting while selecting an appropriate \(k\) that provides sufficient temporal information without excessively complicating the model. This balance will help in achieving both efficient training and robust performance.
\subsection{Generalizable Analysis}
\begin{table}[h]
\centering
\begin{tabular}{c|cc|c|c}
\hline
\multirow{2}{*}{\begin{tabular}[c]{@{}c@{}}Node-Level\\ Method\end{tabular}} & \multicolumn{2}{c|}{Bitcoin-Alpha} & \multirow{2}{*}{\begin{tabular}[c]{@{}c@{}}Edge-Level\\ Method\end{tabular}} & WADI           \\ \cline{2-3} \cline{5-5} 
                                                                             & AUC              & AP              &                                                                              & F1             \\ \hline
GeneralDyG                                                                     & \textbf{96.28}   & \textbf{26.73}  & GeneralDyG                                                                          & \textbf{60.43} \\ \hline
GDN                                                                          & 83.84            & 13.28           & TADDY                                                                        & 40.05          \\
MST-GAT                                                                      & 86.66            & 18.97           & SimpleDyG                                                                    & 33.24          \\
FuSAGNet                                                                     & 87.76            & 20.01           & SAD                                                                   & 36.75          \\ \hline
\end{tabular}
\caption{Generalizable analysis on Node-Level and Edge-Level tasks}
\label{Generalizable}
\end{table}
In Table \ref{Generalizable}, GeneralDyG demonstrates its strong generalizability across different types of tasks. Specifically, GeneralDyG consistently outperforms the baseline methods that were evaluated in a mismatched dataset context. For instance, when the edge-level baselines are applied to the node-level dataset (WADI), their performance significantly drops, with metrics such as AUC and F1 score showing substantial declines compared to GeneralDyG. Similarly, node-level baselines tested on the edge-level dataset (Bitcoin-Alpha) exhibit poor performance, further emphasizing their lack of generalizability.

GeneralDyG, on the other hand, maintains high performance across both types of datasets, showcasing its robustness and adaptability. This indicates that GeneralDyG is capable of effectively handling both node-level and edge-level tasks, whereas the baseline methods exhibit considerable performance degradation when faced with different dataset types. These results underline the superior generalizability of GeneralDyG, as it maintains stable and effective performance across diverse scenarios where other methods fail to deliver consistent results.

\section{Conclusion}
In this work, we introduced a novel approach for anomaly detection in dynamic graphs called GeneralDyG, which effectively addresses the challenges of data diversity, dynamic feature capture, and computational cost, thereby demonstrating the generalizability of our method. GeneralDyG achieves this by mapping node, edge, and topological structure information into the feature space, incorporating hierarchical tokens, and sampling temporal ego-graphs to efficiently capture dynamic features. GeneralDyG excels across multiple benchmarks, demonstrating its effectiveness and high performance. For future work, we can build on this work to explore the interpretability of anomaly detection in dynamic graphs, providing more robust theoretical support.
\appendix

\section{Acknowledgments}
We sincerely thank Wenjie Yin for their insightful suggestions and helpful discussions that greatly contributed to the development of this work. 
This research is supported by the Joint NTU-WeBank Research Centre on Fintech, Nanyang Technological University, Singapore. It is also supported by the Joint NTU-UBC Research Centre of Excellence in Active Living for the Elderly (LILY) and the College of Computing and Data Science (CCDS) at NTU Singapore. This work is partially supported by the Wallenberg Al, Autonomous Systems and Software Program (WASP) funded by the Knut and Alice Wallenberg Foundation. 

% \bigskip
% \noindent Thank you for reading these instructions carefully. We look forward to receiving your electronic files!

\bibliography{aaai25}

\clearpage

\appendix
\section{Appendix}
Tabel \ref{symbol} summarizes the symbols used in the main paper.
\begin{table}[h]
\scalebox{0.92}{\begin{tabular}{cc}
\hline
Symbol                              & Meaning                                                \\ \hline
$\mathcal{G}$                       & continuous-time dynamic graph (CTDG)                   \\
$\mathcal{V}$                       & set of nodes that participate in temporal edges        \\
$\mathcal{E}$                       & chronologically ordered series of edges                \\
$\delta(t)$                         & interaction from node to node in CTDG                  \\
                                    & $\delta(t) = (v_i, v_j, t, e_{ij})$                    \\
$v_i, v_j$                          & $v_i, v_j \in \mathcal{V}$                             \\
$x_{v_i}, x_{v_j}$                  & node attributes for nodes $v_i, v_j$                  \\
$\mathcal{X}$        & node attributes for all nodes                          \\
$e_{ij}$                            & $e_{ij} \in \mathcal{E}$                               \\
$y_{e_{ij}}$                        & edge attributes for edges $e_{ij}$                     \\
$\mathcal{Y}$                       & edge attributes for all edges                          \\
$n$                                 & number of nodes                                        \\
$m$                                 & number of edges                                        \\
$\mathcal{A}$                       & anomaly events                                         \\
$\mathcal{Z}$                       & anomaly features                                       \\
$Q, K, V$                           & query, key, value in Transformer                       \\
$t$                                 & timestamp                                              \\
$f$                                 & learnable anomaly score function                       \\
$f(e)$ or $f(v)$                    & greater likelihood of anomaly for edge $e$ or node $v$ \\
$y_e = 1$                           & anomalous label for edge                               \\
$y_n = 1$                           & anomalous label for node                               \\
$a_i$                               & event and $a_i \in \mathcal{A}$                        \\
$\langle |KHS| \rangle$             & special token                                          \\
$\varphi$                           & GNN extractor                                          \\
$k$                                 & parameter for $k$-hop                                  \\
$\mathcal{K}$                      & parameter for GNN extractor                            \\
$\Bar{A}$                           & Laplacian-adjacency matrix with self-loops             \\
$D$                                 & diagonal degree matrix                                 \\
$I$                                 & identity matrix                                        \\
$\sigma$                            & activation function                                    \\
$T \in \mathbb{R}^{N_v \times N_e}$ & binary transformation matrix                           \\
$\odot$                             & Hadamard product                                       \\
$H^{(K)}$                           & features in the $(K)$-th layer in GNN extractor        \\
$\mathcal{F}{exp}$                  & exponential kernel (non-symmetric)                     \\
W                                   & Weight Matrix used in Transformer                      \\
$W$                                 & Weight Matrix used in GNN extractor                    \\ \hline
\end{tabular}}
\caption{List of symbols used in the paper.}
 \label{symbol}
\end{table}

\subsection{Objective Function}
For anomaly detection on dynamic graphs, we define an objective function for node-level and edge-level tasks using binary cross-entropy loss. For a node $v \in V$ with label $y_n$ (1 for anomalous, 0 for normal) and score $f(v)$, and an edge $e \in E$ with label $y_e$ and score $f(e)$, the objective function is:

\begin{align}
\mathcal{L} = & \, \alpha \left(- \sum_{v \in V} \left( y_n \log(f(v)) + (1 - y_n) \log(1 - f(v)) \right)\right) \notag \\
& + \beta \left(- \sum_{e \in E} \left( y_e \log(f(e)) + (1 - y_e) \log(1 - f(e)) \right)\right)
\end{align}

where $\alpha$ and $\beta$ are indicators for including node-level and edge-level tasks, respectively.

\subsection{Pseudo Code}
\begin{itemize}[left=0pt]
    \item \textbf{Input:} Dynamic graph \( G = (V, E) \), anomaly event sequence \( a_i \)
    \item \textbf{Output:} Anomaly scores \( f(v) \) for nodes \( v \) and \( f(e) \) for edges \( e \)
    \item \textbf{Temporal Ego-Graph Sampling}
    \begin{itemize}
        \item For each event \( a_i \):
        \begin{itemize}
            \item Extract \( k \)-hop ego-graph around \( a_i \)
            \item Form sequence \( w_i \) with temporal ordering and add special tokens
        \end{itemize}
    \end{itemize}
    \item \textbf{Temporal Ego-Graph GNN Extraction}
    \begin{itemize}
        \item For each event \( a_i \):
        \begin{itemize}
            \item Compute structural embedding \( \phi(z_i) = \text{GNN}(w_i) \) using TensGNN
        \end{itemize}
    \end{itemize}
    \item \textbf{Temporal-Aware Transformer}
    \begin{itemize}
        \item Compute attention embeddings \( \text{Attn}(z_i) \) for each \( z_i \) and Scoring
    \end{itemize}
\end{itemize}

\subsection{Supplementary for Experiments}
\subsubsection{Baselines}
\begin{itemize}
    \item \textbf{node2vec} performs random walks on the graph to generate sequences of nodes, which are then trained using the skip-gram model, embedding nodes into a continuous vector space while preserving structural information and node similarities.
    \item \textbf{DeepWalk} combines random walks with the skip-gram model to learn latent representations of nodes, capturing the community structure and node similarities in a continuous vector space.
    \item \textbf{TGAT} uses attention mechanisms and temporal encoding to capture the evolving structure and features of nodes over time for inductive representation learning on dynamic graphs.
    \item \textbf{TGN} incorporates memory modules to store historical information and uses message passing to update node embeddings as new events occur over time in dynamic graphs.
    \item \textbf{StrGNN} leverages the structural properties of graphs, using graph neural networks to capture both local and global patterns for identifying anomalies.
    \item \textbf{TADDY} models the temporal dependencies and evolving structures within dynamic graphs to identify unusual patterns and detect anomalies.
    \item \textbf{SAD} utilizes a time-equipped memory bank and a pseudo-label contrastive learning module to effectively leverage both labeled and unlabeled data, enabling the discovery of underlying anomalies in evolving graph streams.
    \item \textbf{SLADE} monitors changes in node interaction patterns over time in dynamic edge streams, using self-supervised tasks to identify abnormal states without relying on labeled data.
    \item \textbf{PCA} reduces the dimensionality of data while retaining the most significant variance, allowing for the identification of outliers based on their deviation from the principal components.
    \item \textbf{KNN} identifies data points that have few or distant neighbors, indicating potential outliers compared to the rest of the data.
    \item \textbf{GDN} leverages graph neural networks to learn the normal patterns in graph-structured data, identifying anomalies by measuring deviations from these learned patterns.
    \item \textbf{BTAD} uses a Bi-Transformer structure and an adaptive multi-head attention mechanism to extract features and analyze trends efficiently for detecting anomalies in multivariate time series data.
    \item \textbf{GRN-100} utilizes a gated recurrent network (GRN) to model temporal dependencies in time series data, identifying anomalies by capturing long-term patterns and deviations.
    \item \textbf{DAGMM} combines a deep autoencoder with a Gaussian Mixture Model (GMM) to jointly optimize data reconstruction and density estimation, detecting anomalies in high-dimensional data without pre-training.
    \item \textbf{MST-GAT} uses a multimodal graph attention network and a temporal convolution network to capture spatial-temporal correlations in multimodal time series, improving detection accuracy and interpretability by explicitly modeling inter- and intra-modal relationships.
    \item \textbf{FuSAGNet} combines Sparse Autoencoder and Graph Neural Network to jointly optimize reconstruction and forecasting while explicitly modeling relationships in multivariate time series.
    \item \textbf{LSTM-VAE} fuses high-dimensional, multimodal sensory signals to detect anomalies by capturing temporal dependencies with LSTM and reconstructing expected distributions with a variational autoencoder.
    \item \textbf{MTAD-GAT} employs graph attention layers to capture dependencies in both temporal and feature dimensions, jointly optimizing forecasting and reconstruction models to enhance detection accuracy and interpretability in multivariate time series.
\end{itemize}
\subsubsection{Experiment Description}
In this work, the parameters for GeneralDyG are set as $k$=2 and $\mathcal{K}$=2. We recommend not setting $k$ greater than 3 for dynamic graph anomaly detection tasks. This is because most existing dynamic graph anomaly detection datasets are large, and when $k>$3, the size of the extracted ego-graph far exceeds that of ego-graphs with $k$=1 or $k$=2, without providing significant performance improvement. Therefore, setting $k$ to 1 or 2 is the most appropriate. Additionally, we did not place excessive emphasis on other parameters. The hidden size of the GNN extractor is set to 128, the hidden size of the Transformer is set to 258, the number of heads in the Transformer is set to 4, and the number of layers is set to 6.
\subsection{Impact of Special Tokens}
\begin{table}[h]
\centering
\scalebox{1}{
\begin{tabular}{c|ccc}
\hline
\multirow{2}{*}{Method} & \multicolumn{2}{c}{Bitcoin-Alpha} & WADI           \\ \cline{2-4} 
                        & AUC             & AP              & F1             \\ \hline
GeneralDyG                    & \textbf{96.28}  & {\underline{ 26.73}}     & \textbf{60.43} \\ \hline
no special              & {\underline{ 96.27}}     & \textbf{26.80}  & {\underline{ 60.11}}    \\
no sort                 & 95.37           & 26.09           & 59.74          \\
no sort \& special      & 95.20           & 26.11           & 59.70          \\ \hline
\end{tabular}}
\caption{Impact of Special Token and Temporal Sorting on Temporal ego-graph sampling}
\label{special}
\end{table}
Table \ref{special} evaluates the impact of special tokens and temporal sorting on the performance of the $k$-hop ego-graph model. The results indicate:
\begin{itemize}
    \item \textbf{No Special Token:} Removing the special token (no special) results in a minor decrease in performance, with AUC and F1 scores slightly lower than those of GeneralDyG. This suggests that the hierarchical information provided by the special token may be adequately captured by the TensGNN itself.
    
    \item \textbf{No Temporal Sorting:} Omitting temporal sorting (no sort) leads to a substantial drop in both AUC and F1 scores. This significant reduction highlights the importance of temporal sorting in preserving temporal dependencies and improving anomaly detection accuracy.
    
    \item \textbf{No Special Token and No Temporal Sorting:} The combination of removing both the special token and temporal sorting (no sort \& special) results in the worst performance among the variants. This further emphasizes the critical role of both temporal sorting and special tokens in achieving optimal model performance.
\end{itemize}
Overall, while the special token has a relatively minor impact, temporal sorting is crucial for maintaining high performance. The combination of both features is essential for effectively capturing the temporal and hierarchical structures in the data.

\subsection{Ablation experiment supplement}
We supplemented ablation experiments on Bitcoin-Alpha with 1\% and 5\% anomaly ratios. The AP results are summarized in the table below:

\begin{table}[htbp]
\centering
\caption{Ablation Study Results on Bitcoin-Alpha (AP values)}
\begin{tabular}{lcc}
\toprule
\textbf{Method}           & \textbf{1\%} & \textbf{5\%} \\
\midrule
GeneralDyG (Full)         & 24.00                      & 24.02                      \\
W/o ego-graph sampling    & 14.72                      & 18.69                      \\
W/o TensGNN               & 17.99                      & 19.90                      \\
W/o Transformer           & 21.78                      & 17.24                      \\
\bottomrule
\end{tabular}
\end{table}

These results demonstrate that Temporal ego-graph sampling significantly improves the AP under different anomaly ratios.

\subsection{Supplement for effect of parameters $k$ and $\mathcal{K}$}

We can observe that as Temporal ego-graph sampling gradually loses effectiveness (with increasing values of $k$ and $\mathcal{K}$), the AUC of the prediction results gradually decreases. For the final version, we added supplementary experiments to investigate the variation in AP under the same conditions. The results are summarized in the table below:

\begin{table}[htbp]
\centering
\caption{AP Results under Different $k$ and $\mathcal{K}$ Values}
\begin{tabular}{cccccc}
\toprule
\textbf{$k$} & \textbf{$\mathcal{K}=1$} & \textbf{$\mathcal{K}=2$} & \textbf{$\mathcal{K}=3$} & \textbf{$\mathcal{K}=4$} & \textbf{$\mathcal{K}=5$} \\
\midrule
1 & 24.12 & 26.34 & 25.80 & 24.57 & 22.91 \\
2 & 25.03 & 26.73 (peak) & 25.63 & 23.99 & 22.77 \\
3 & 24.01 & 24.45 & 24.21 & 23.38 & 23.01 \\
4 & 23.21 & 23.62 & 22.95 & 22.84 & 21.45 \\
5 & 22.89 & 22.57 & 21.88 & 21.62 & 21.11 \\
\bottomrule
\end{tabular}
\end{table}

These results show that Temporal ego-graph sampling is a necessary step to ensure good AP.
\end{document}